# Unmanned Surface Vehicle Path Planning from the Perspective of Multi-Modality Constraints: A Comprehensive Analysis

Chunhui Zhou [a,b,c], Shangding Gu [a,b*], Yuanqiao Wen [c,d*], Zhe Du [e], Changshi Xiao[a,b,c], Liang Huang [b,c,d], Man Zhu [b,c,d]

(a. School of Navigation, Wuhan University of Technology, Wuhan, 430063, China.

b. Hubei Key Laboratory of Inland Shipping Technology, Wuhan, 430063, China.

c. National Engineering Research Center for Water Transport Safety, Wuhan, 430063, China.

d. Intelligent Transport Systems Center, Wuhan University of Technology, Wuhan, 430063, China.

e. Department of Marine and Transport Technology, Faculty of Mechanical, Maritime and Materials Engineering, Delft University of Technology, Delft, 2628BX, The Netherlands.)

**Abstract:** The essence of the path planning problems is multi-modality constraint. However, most of the current literature has not mentioned this issue. This paper introduces the research progress of path planning based on the multi-modality constraint. The path planning of multi-modality constraint research can be classified into three stages in terms of its basic ingredients (such as shape, kinematics and dynamics et al.): Route Planning, Trajectory Planning and Motion Planning. It then reviews the research methods and classical algorithms, especially those applied to the Unmanned Surface Vehicle (USV) in every stage. Finally, the paper points out some existing problems in every stage and suggestions for future research.

* Corresponding Author. E-mail address: gshangd@163.com

* Corresponding Author. E-mail address: 3444324@qq.com

# 1. INTRODUCTION.

## 1.1 BACKGROUND INTRODUCTION

With the development and application of artificial intelligence and machine learning, more and more studies focus on unmanned vehicles and their applications (Zhou, Z., 2016). For example, Unmanned Ground Vehicle (UGV) or wheeled robot is widely used in field of industrial automation (automatic forklift), warehouse management, planet exploring (lunar rover), disaster rescue, intelligent transportation (automatic drive) and military operation (de-mining robot) (Arai et al., 2002; Farinelli et al., 2004; Kui et al., 2007). The application of Unmanned Aerial Vehicle (UAV) is also increasingly changed from military domain to civil use, such as remote sensing photographing, agricultural spraying, communications relay, environmental monitoring and express service (Jayoung et al., 2013; George et al., 2012; Mingzhu et al., 2016). The development of UGV and UAV has already been updated to a new level.

Another unmanned vehicle should also be paid attention to, which is the Unmanned Surface Vehicle (USV). The application scenarios are not widely applied for civil use and the studies of a USV are relatively fewer and commence a bit late. However, USVs have played an important role in some areas like ocean sampling, seabed mapping, network monitoring cooperating with UAV and UGV, search and rescue, and harbor patrolling (Shafer, et al., 2008; Simetti, et al., 2009; Steimle, E. T. & Hall, M. L., 2006; Vitor A. M. Jorge et al., 2019; Liu, Z. et al., 2016). In addition, USVs also have great potential in urban waterways cargo transport for relieving the congestion in the overloaded road networks (Chen, L., et al., 2019). Nowadays, many countries become more and more interested in rich maritime resources and some of them have already cataloged this into their national development strategy. The USV is an important tool for exploring oceans (Fossen, T.I. 1994; Liu, Z. et al., 2016). Thus it should be well studied.

No matter what type the unmanned vehicles are, a vital character for all of them is autonomy. The word "*autonomy*" is originated from the ancient Greece, which consists of "*αυτο*" (means "self") and "*νομοζ*" (means "law") (Lekkas, 2014). The autonomy of unmanned vehicles is the ability that can make a plan according to some rules (laws) and execute it all by themselves in a certain environment in the absence of human intervention. Especially, a very important part of autonomy is to work out a safe path for the vehicle, also called Path Planning (Murphy, R. 2000; H. Choset, 2005). Path planning is one of the key technologies in the process of automating and intelligentizing USVs and carrying out complex tasks of USVs (Fossen, T.I. 2002; Lazarowska A.,

2015; Song, R. et al., 2017).

When carrying out researches on Path Planning, many researchers always cite Durrant-Whyte Hugh's three questions to describe the issue, "Where am I?", "Where am I going?" and "How do I get there?" (Durrant-Whyte Hugh, 1994). These questions are applicable to an early stage of Path Planning, because at this stage both the planning environment (like indoor or virtual environment) and the research objects (like game character or wheeled robot) were relatively simple. When it comes to a UGV and a UAV, due to the circumstances either fixed (the urban road network) or ample (the expansive airspace), and relatively easy control (they can be controlled positioning, like braking and hovering), the three questions are appropriate.

For the USV, the environment is far more complex. Apart from the interference of wind, wave and current, the USV's inertia, in some environments the resistance and the response time on waters are also larger than those on the ground and in the air (Fossen, T.I. et al., 2017; Zhang, F. et al., 2015). The control of the USV is harder than the UGV and the UAV to some extent. Thus, the method on traditional path planning and these three questions might be not applicable to the USV. In fact, the path planning of the USV requires more description, such as the USV's situation, response time and inertia factor, etc. And the "old" three questions should be changed to new four questions, "**What's my situation?**", "**Where can I go?**", "**Where should I go?**" and "**How should I do to get there**" (Table 1 shows the contrast and analysis of UGVs, UAVs and USVs, Table 2 shows the USVs research For various applications). In addition, we referenced the literature (Jin K. F., et al., 2018) that illustrated the 5 levels of USVs automation and intelligent. Further, we expanded and refined the USVs automation and intelligent level classification (Table 3 shows the USVs automation and intelligent level classification).

Table 1. UGVs, UAVs and USVs Contrast and Analysis

| Type / elements | UAVs | UGVs | USVs |
| --- | --- | --- | --- |
| Motion speed & inertia | Fast motion and small inertia | Fast motion and small inertia | Fast motion and great inertia |
| Major disturbing factors | Wind | Pavement | Wind, current and wave |
| Running scenarios | The expansive airspace | The road network | The waters environment |
| Response time | Small | Small | Large |
| Application scenarios | Remote sensing photographing, | Industrial automation, | Ocean sampling, seabed mapping, |
| | agricultural spraying, communications relay, environmental monitoring and express service, etc. | warehouse management, planet exploring, disaster rescue, intelligent transportation and military operation, etc. | network monitoring cooperating with UAVs and submarines, search and rescue, and harbor patrolling, etc. |

Table 2 USVs of Various Applications

| USV use division | Specific applications type |
|---|---|
| Water sampling and monitoring | USV for water quality monitoring and measurement (T. H. Yang et al., 2018)， USV for shore monitoring (G. Hitz et al., 2016), USVs as environmental monitoring and assessment tools (E.T., 2006; W. Naeem et al., 2008), Key technology of water sample collection for USV (F. Li, 2016). |
| Antisubmarine | USV anti-submarine and combat mode (W. Z. Li et al., 2018), USV antisubmarine and load analysis (J. T. Su, 2018), Unmanned vehicles for anti-submarine warfare (Fahey, S., & Luqi. 2016). |
| Target tracking | Time-space fusion of USV target tracking (Y. Peng et al., 2018), USV target detection and tracking based on ssd-cf (X. J. Chen et al., 2019), Target path tracking control optimization for USV (X. Y. Cong et al., 2019), Target detection and tracking of USV based on optical vision (W. J. Zeng, 2013). |
| Maritime search and rescue | USV water search and rescue (A. J. Shafer et al., 2008; G. A. Wilde and R. R. Murphy 2018), Power design for search and rescue USV (J. Wang et al., 2017), The direction and speed control of search and rescue USV (X. D. Tan, 2019). |
| Patrol USV | Patrol USV based on metacentric automatic adjustment function (S. Zhao et al., 2018), USV information collection and patrol (G. Oriolo et al., 1995), USVs operating in harbour fields (Casalino, G. et al., 2009); inspect littoral structures (Steimle, E., et al., 2009); complete coverage path planning of USV for island mapping (Yuxuan, Z., et al., 2017) |

Table 3 USVs automation and intelligent level classification

| Classification | Level Description | Current Research Status |
| --- | --- | --- |
| Level one | Remotely control and plan the ship's path. | have achieved (Caccia, M. et al., 2008) |
| Level two | The ability of path tracking. | have achieved (W. J. Zeng, 2013; X. J. Chen et al., 2019) |
| Level three | Optimal path planning capability under static obstacles. | have achieved (Thakur, A. er al., 2011; Blaich, M. et al., 2012; Song, R. et al., 2019; Xiaojie Sun, 2016; Lei X. et al., 2019;) |
| Level four | Under dynamic obstacles, better path planning is achieved. | have achieved (Jing, L. et al., 2015; Huang Y. et al., 2018a&b; Shi, B. et al., 2019) |
| Level five | Under complex environment (especially wind, current, wave and dynamic obstacles), optimal path planning. | Partial functions have achieved (Yoo, B., & Kim, J., 2016; Sarda, E. I., et al., 2016; Song, R., et al., 2017; Ma, Y., et al., 2018; Singh, Y. et al., 2018; Wang, N., et al., 2019) |
| Level six | Realize autonomous formation and cooperative control path planning. | Partial functions have achieved (Ihle, I.A.F. et al., 2007; Chen, L., et al., 2019; Ali, H. and Rudy, R.N. 2019; Shijie, L. et al., 2019) |
| Level seven | Realizing autonomous navigation and cognitive planning in a complex environment. | Not yet achieved |
| Level eight | Realization of group USVs game and autonomous navigation in complex environment. | Not yet achieved |

## 1.2 THE MAIN WORK AND CONTRIBUTION OF THE ARTICLE

This paper reviews the researches of path planning based on multi-modality constraint for the USV. Firstly, the research progress of path planning is introduced. Then following a clue of the research progress, the related research methods and classical algorithms are reviewed, especially those applied to the USV, which mainly sorts out the researches related to the USV path planning in the past two decades, analyzes the USV's path planning constraint elements, and divides the USV path planning into three stages from the perspective of multi-modality constraint, and each stage of path planning is analyzed and discussed, which provides a certain reference for the future research in this field (due to the limitation of space, some excellent USVs' path-planning literature are not mentioned in the text, for which we are deeply sorry).

## 1.3 THE MAIN STRUCTURE OF THE ARTICLE

Section 1 of the article introduces the USV's feature and summarizes the whole text. Section 2 mainly analyzes the constraint elements of different modality: planning space, planning time, planning behavior and planning criterion; Sections 3 to 5 divide the USV path planning into three stages based on the modality-constraints elements of the second part: route planning, trajectory planning and motion planning, and analyze the progress and problems of each stage. Section 6 mainly summarizes the whole article and looks forward to the future development of the USV's path planning.

# 2. THE PROGRESS OF PATH PLANNING.

Steven M. LaValle (Steven, 2006) mentioned in his book "*Planning Algorithm*" that planning is of different meanings in different disciplines. In the field of Robotics, it focuses on the automated mechanical systems which are capable of sensing, driving and computing. The purpose of the planning is to transform the specific tasks that the robot needs to perform (high-dimensional) into the manipulation of the robot (low-dimensional) (Lydia, et al., 2008). In the field of USV Control Theory, one part is to design inputs to physical systems described by differential equations; Another part focuses on the optimal control, namely on the premise of the minimum consumption of the resources (such as time and energy saving), control the system running. And in the field of Artificial Intelligence, the researchers tend to discretize the problem (H. Choset, et al., 2005), which solve the problems like the Rubik's cube or a sliding-tile puzzle that needs to build a discrete model, and generally planning time is discrete.

There are several basic ingredients that the USV planning requires: Planning Space, Planning Time, Planning Behavior and Planning Criterion (E. Plaku, et al., 2005).

Planning Space not only refers to the geographic and environmental information, but also contains the dynamic performance of the research objects and other state-space information. Namely, it covers the necessary information during the planning process. The information is fundamental to the design of the planning algorithm. The information of the initial and goal state is particularly important because the planning algorithm is usually a rule that describes the process from the initial state to the goal state.

Planning Time refers to the sequence of decisions that must be made over time. In general, Planning-Time issues not only focus on completing a task within a specific time but more importantly, the planning process must be followed by the correct and reasonable order of execution (i.e., step by step).

Planning Behavior is the actions through which the USVs could reach the goal state from the initial state. In some cases, there are no restrictions on those actions, and the process of planning is ideal. But for most of the time, the actions are restrained by the natural environment and the dynamic performance of the research objects.

Planning Criterion is the desired constraint condition based on Planning Space and Planning Behavior. There are generally two different types of criterion: feasibility and optimality. Feasibility is to find a plan by which the USV could reach the goal state. Optimality is to find an efficient plan for the shortest voyage, the shortest time-consuming, the lowest fuel-consuming or the hybrid optimization.

USV Path planning problem is a typical problem in planning issues. Path planning problem can be defined as a planning problem with the intention of planning out a path which meets certain requirements. Different modality requirements mean different Planning Criteria (modality refers to a mode of planning). Planning Criterion is related to Planning Space, Planning Time, Planning Behavior and Planning Criterion. So according to the different modality constraints of the basic ingredients, the development of USV path planning has been divided into three stages: Route Planning, Trajectory Planning and Motion Planning.

The progress of Path Planning based on multi-modality constraint is shown in Figure 1.

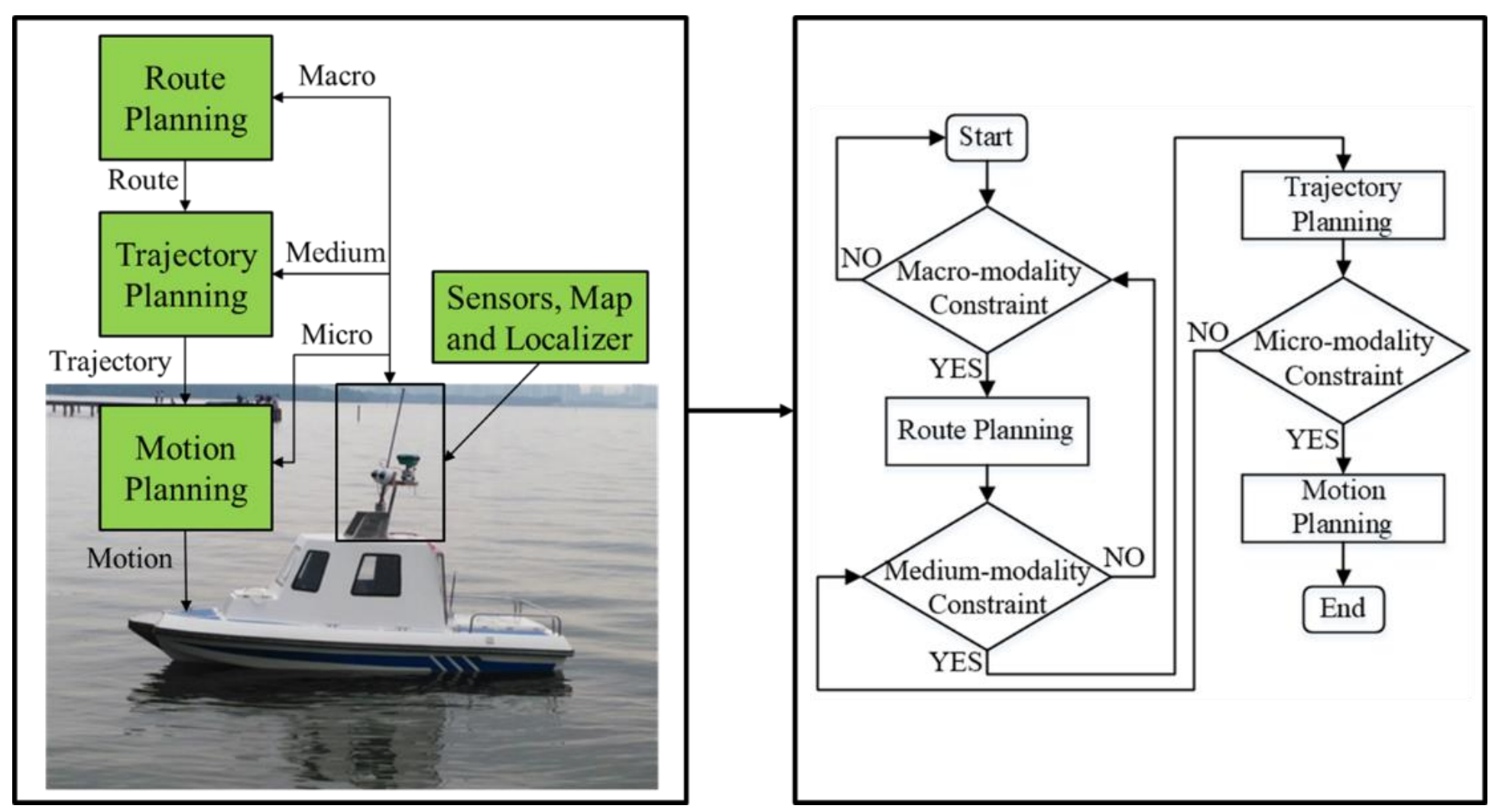

Figure 1 USV Path Planning based on multi-modality constraint

USV Route Planning is an initial stage and is defined as a macro-modality constraint problem. In this stage, the USV is taken as a particle ignoring its scale character and dynamic performance. Planning Space mainly considers the constraints of geography and Planning Behavior is not constrained by the actions of the research object. Planning Criterion focuses on the quality of the path itself. The desired outcome is an optimal route from the starting point to the end point.

USV Trajectory Planning is a transition stage and is defined as a medium-modality constraint problem. In this stage, the Trajectory Planning object takes part of the dynamic constraints into account (such as the shape and kinematics, etc.) and it considers environment constraints of the research object in Planning Space (such as the channel width). Planning Behavior is restricted by the influence of certain part of these constraints (such as linear speed and angular velocity, etc.). Planning Criterion is the improvement for the planned route of the last stage. So the desired outcome in this stage is the optimal trajectory from the starting point to the end point.

USV Motion planning is a final stage and is defined as a micro-modality constraint problem. In this stage, the Motion Planning object is considered with more comprehensive dynamic constraints, including the constraints in Planning Space and Planning Behavior, and it combines path planning with motion control. The desired outcome is a practical motion path from the starting point to the end point.

The ingredients of Path Planning based on multi-modality constraint are shown in Table 4.

Table 4. The ingredients of Path Planning based on multi-modality constraint

| Development / Ingredients | USV Route Planning | USV Trajectory Planning | USV Motion Planning |
| --- | --- | --- | --- |

| Planning Space | Environment Conditions | Environment Conditions + Part (Kinematics) Constraints | Environment Conditions + fully Dynamic (kinematics +dynamics) Constraints |
| --- | --- | --- | --- |
| Planning Time | Sequence of Decisions (Discretization) | Sequence of Decisions (Discretization) | Sequence of Decisions (Discretization) |
| Planning Behavior | No Restrictions on Research Objects’ Actions | Partial Restrictions on Research Objects’ Actions | Full Restrictions on Research Objects’ Actions |
| Planning Criterion | Plan Out an Optimal Route | Plan Out an Optimal Trajectory | Plan Out an Optimal Motion Path |

In term of routing planning, trajectory planning and motion planning, the main difference is that the specific constraint factors are variational. In other words, the specific application scenarios are different.

As for routing planning like the ship’s path planning in the large scale area, for example, the routing planning for a ship which will navigate from Hawaii port to Panama port, it does not need to consider full detail about how the ship will reach Panama port. In this stage, we should regard the ship as a particle and ignore other factors such as the specific ship’s shape and ship’s speed and so on. Figure 2 shows a ship routing planning from Hawaii port to Panama port.

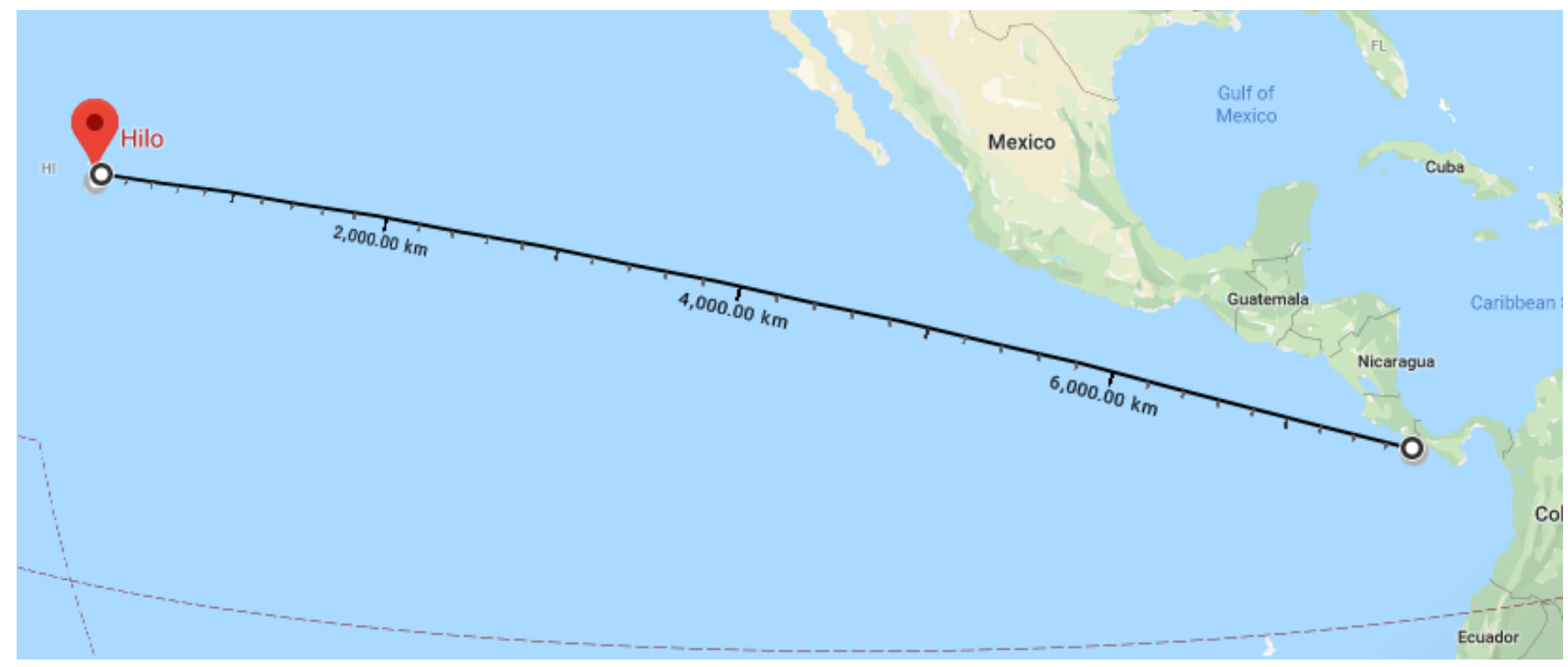

Figure 2 a ship routing planning from Hawaii port to Panama port.

As for trajectory planning like the ship’s path planning in the middle scale area, for instance, when the ship reaches Panama port, it needs to think how to enter the inner port, it needs to consider some constraint factors such as specific ship’s size and speed and so on. In this stage, we should consider some constraints and ignore some unnecessary factors such as moments and forces and so on. Figure 3 shows a ship trajectory planning for how to enter the inner port, the ship is navigated under trajectory planning, in which there are some information that the ship needs to know, such as which entrance is suitable to enter according to the ship sizes, the entrance sizes and its kinematics (e.g. velocity and accelerated velocity etc.).

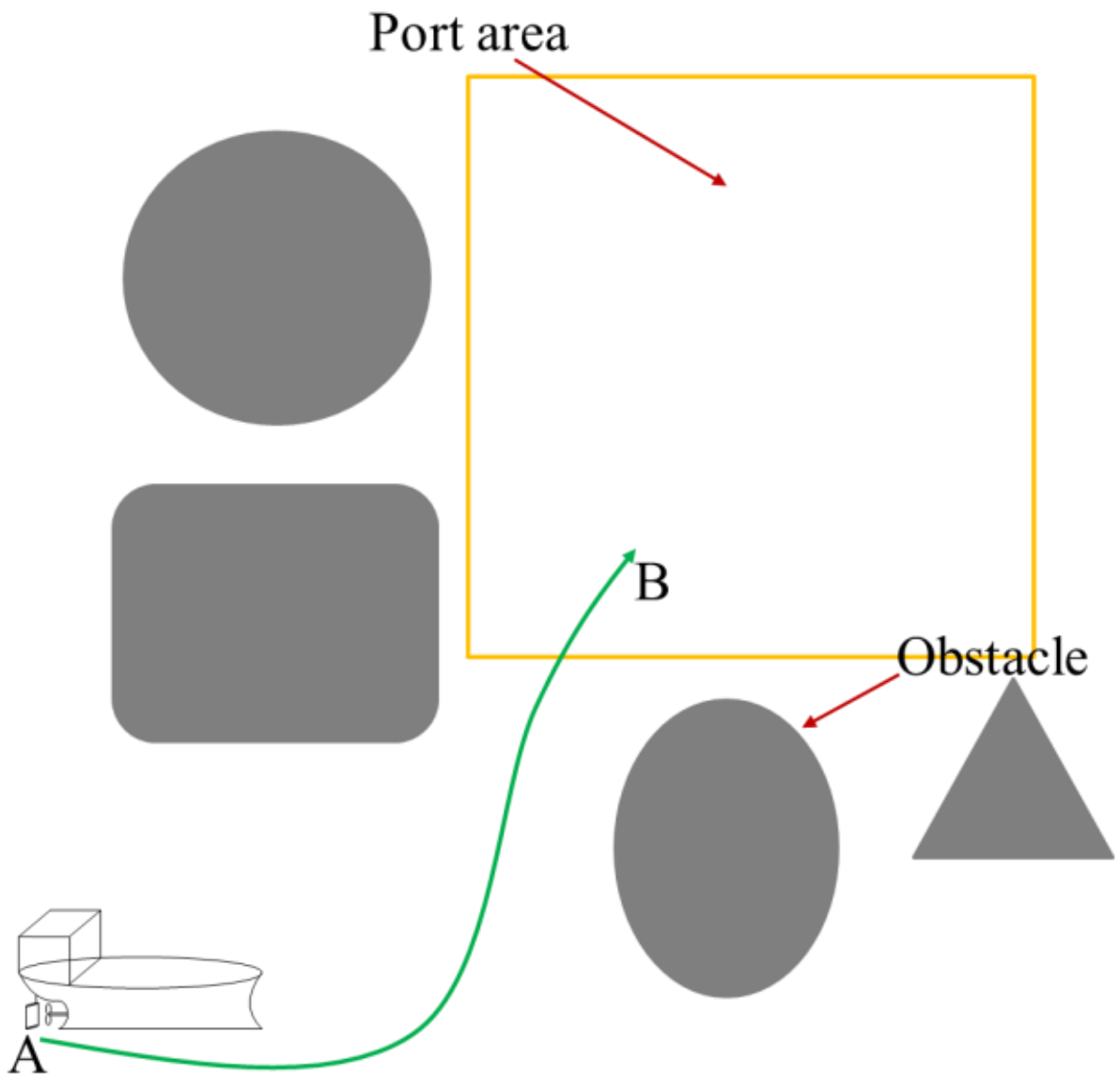

Figure 3 a ship trajectory planning for how to enter the inner port.

As for motion planning like the ship's precise path planning in the small scale area, for example, the motion planning is that a ship berths alongside, which needs to know how to precisely steer to berth. In this stage, the ship is regarded as a rigid body, and we should consider fully constraint factors such as the ship's force and inertia moment and so on. Therefore, the ship's motion planning is applied to more scenarios that require precise tasks such as berth alongside and unberthing operation, etc. Figure 4 shows a ship motion planning for how to berth, the direction of ship at point A is north 0 degree (the blue arrow at point A), the ship needs to know how to steer to reach point B, and keep its direction along the shoreside (blue area) and maintain the direction of blue arrow at point B, in which the ship needs to consider its dynamics (such as inertia forces and moments of the ship's surging, swaying and yawing etc.) and kinematics (such as ship's velocity) and its specific sizes constraints.

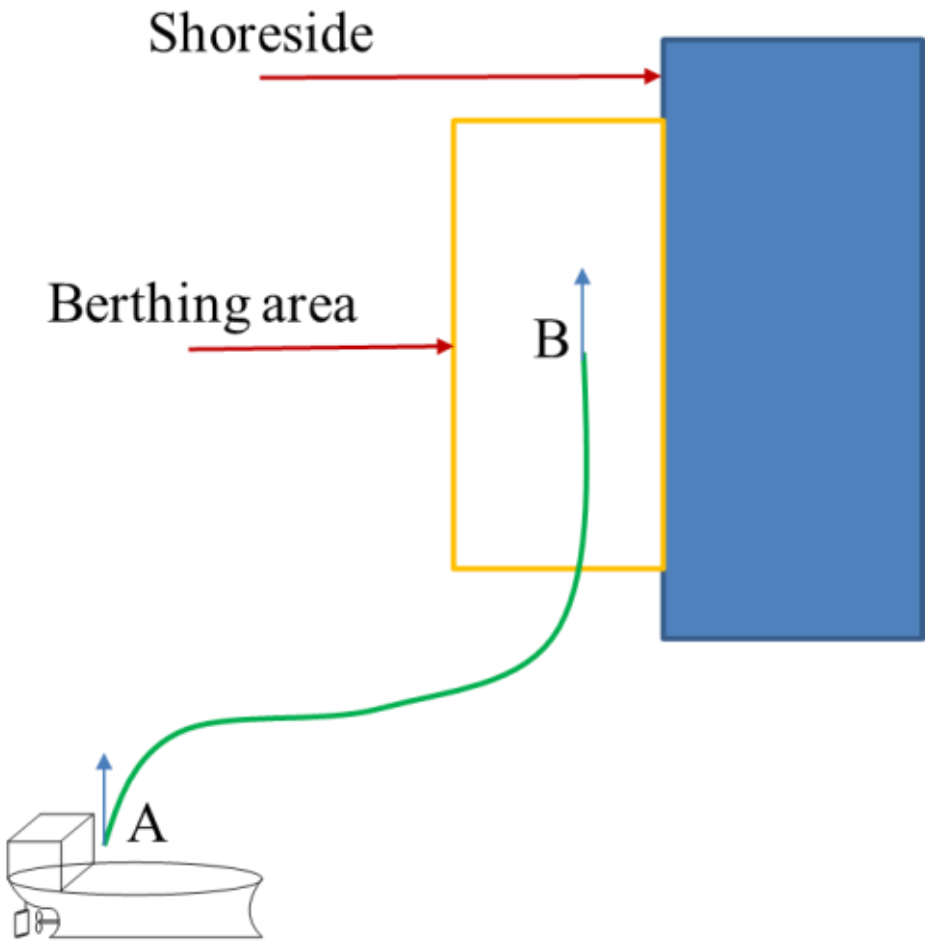

Figure 4 a ship motion planning for how to berth.

Overall, of all the ingredients of USV Path Planning based on the multi-modality constraint, the purpose of the research is to improve the efficiency and effectiveness of the Path Planning. The next three chapters will give specific details of the application scenarios and typical methods for the three stages, respectively. In particular, methods for USV in these three stages are reviewed in detail.

# 3. ROUTE PLANNING.

At the basic stage, Route Planning is defined as a macro-modality constraint problem, that is, its application scenarios are often under large scale environments. In such circumstances, the USV is usually taken as a particle and its dynamic performance is ignored. Route Planning is usually applied to the macro-modality constraint environment, such as the environment shown in Figure 5, in which we do not consider its specific sizes constraints and any dynamic constraints.

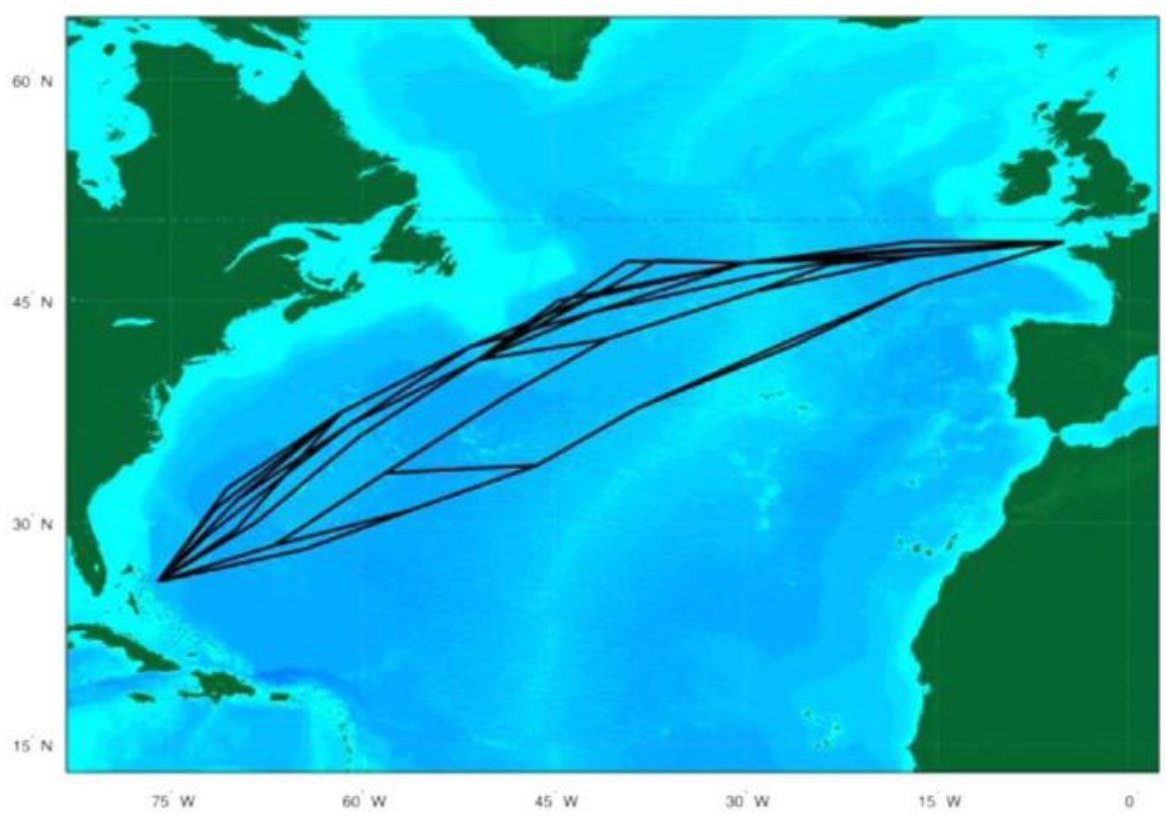

Figure 5 the application in USV Route Planning (Vettor, R., & Guedes Soares, C.. 2016)

# 3.1 TWO IMPORTANT PARTS OF ROUTE PLANNING

Route Planning has been studied for many years and many classical algorithms have been proposed. Most of these algorithms take two steps: environment modeling and optimal path searching.

## 3.1.1 ENVIRONMENT MODELING

Environment modeling is the foundation for Route Planning. In this step, not only the planning space is transformed to a geometry space, but also the whole space is discretized into multiple sub-spaces. Visibility Graph, Voronoi Diagram, and grid Map are the common modeling methods (Latip, N. B. A. et al., 2017; Kim, J. et al., 2011; Huajun, L. et al., 2006; Huai Yang. 2016; Chao C., et al., 2014; Candeloro M. et al., 2017; Niu et al., 2019; Yang, J. M. et al., 2015). One of the most widely used ways of environment modeling in Route Planning is Grid Map (Raster Map). This approach decomposes a planning space into many rectangular areas, usually called grids. Figure 6 shows the Visibility Graph, Figure 7 shows the Voronoi Diagram, Figure 8 shows the grid map.

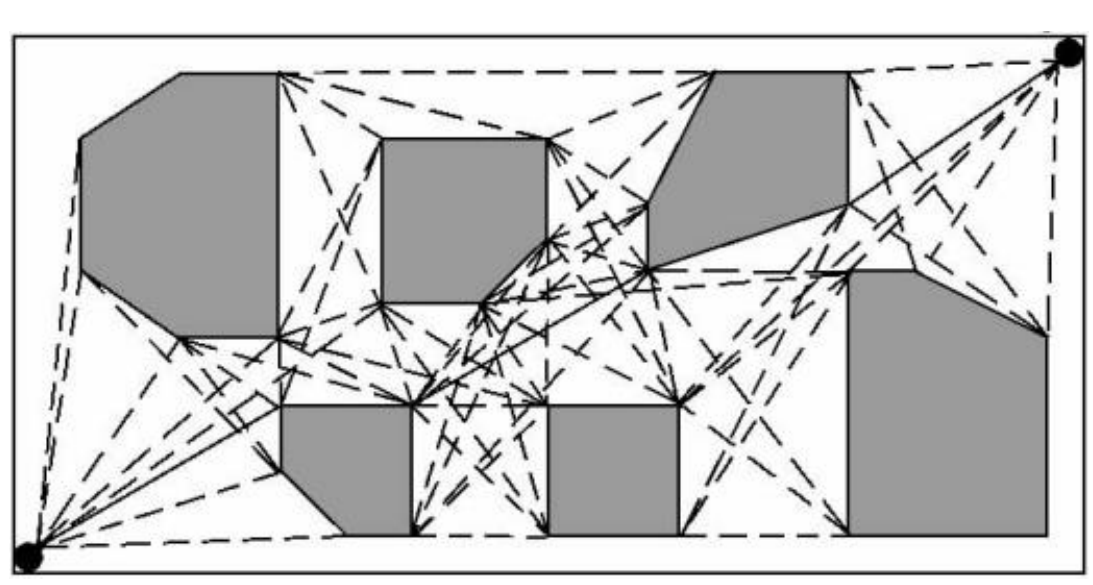

Figure 6 Visibility Graph (Chao C., et al., 2014)

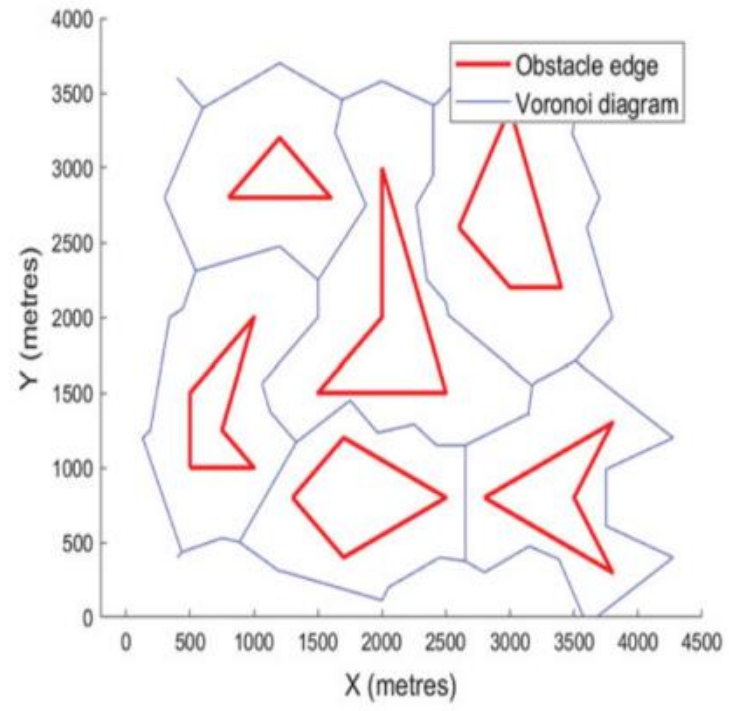

Figure 7 Voronoi roadmap of polygon obstacles (Niu, H., et al., 2019)

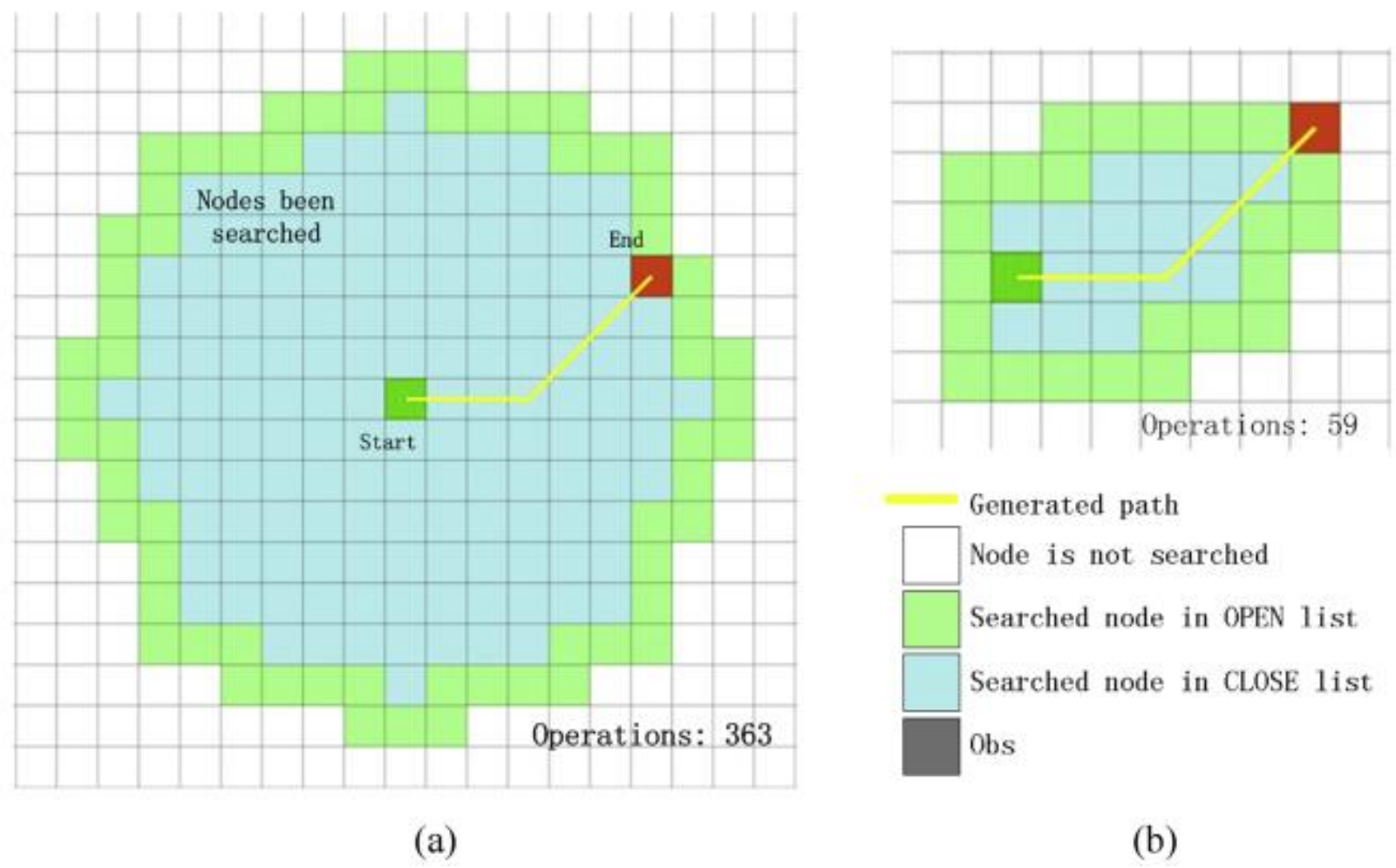

Figure 8 the Grid Map (Song, R. et al., 2019)

Generally, search-path time, search-path completeness and search-path optimality are the main factors for performance of search-path algorithm (Latip, N. B. A. et al., 2017). The three environment models for performance of search-path algorithm have been discussed according to the above three factors.

(1) search-path time.

As for Visibility Graph method, search-path time of visibility graph method mostly depends on the obstacles, which means search-path time will change according to change of number of obstacles, specifically, it will increase if obstacles of search area change.

As for Voronoi Diagram, generally, the complexity of search-path time can be considered as $O(nlog(n))$, in which n is the number of the obstacles vertices (Niu et al., 2019), and for Visibility graph method, the complexity of search-path time can be considered as $O(n^2)$ (Ghosh and Mount, 1991).

As for Grid Map, the search-path time will depend on the density of grid map resolution. If the distance between the grid lines is too large, many environment information will not be shown, and an accurate path search cannot be performed. If the distance between the grid lines is too small, and a small distance between the grid lines will lead to the enormous search-path time, meanwhile if the number of grids increase, the searching space might grow enormously (Lei X. et al., 2019).

(2) search-path completeness.

As for Visibility Graph method, the search-path completeness of Visibility Graph method is better compared to Voronoi Diagram and Grid Map, because paths are selected according to the relationship between obstacles. For Voronoi Diagram, because it is expanded by obstacles, there are fewer search nodes, and the number of search nodes and search freedom is less than that of Visibility Graph. For Grid Graph, it is mainly based on the direction of grid to select the path, the path selections are 8 or 16 directions, so the selection of path points will be more constrained than Visibility Graph and Voronoi Diagram (Zhe Du et al 2018 & 2019).

(3) search-path optimality.

Generally, for the Visibility Graph, the search path is relatively short, but this path often does not meet the practical requirements, and further optimization is needed. For the Voronoi Diagram, the search path is generally longer than that of the Visibility Graph, but the search path is often farther from the obstacles than that of the Visibility Graph, that is to say, the search path is longer than that of the Visibility Graph, and the path that the Voronoi Diagram get is more secure. For Grid Map, due to the jag effect of Grid Map, the search path of Grid Map is not optimal, but in recent years, there are many scholars that smooth the grid-nodes search, that is to delete unnecessary search nodes to make it smoother and shorter path (Hongwei, W., et al., 2010; Song, R. et al., 2019).

### 3.1.2 OPTIMAL PATH SEARCHING

Optimal path searching algorithms play an important role in macro-modality constraint Route Planning.

Searching algorithms are mainly divided into the heuristic method, evolutionary algorithm (swarm intelligence algorithm) and potential field algorithm. Heuristic algorithms include A* algorithm (Hart, P. et al., 1968) and Lifelong planning A∗ (Koenig, S., et al., 2004), etc; Evolutionary algorithms include Particle Swarm Optimization (PSO) algorithm (Shi, Y., & Eberhart, R. C. 2002; Ma, Y et al., 2018; Liang et al., 2019), Genetic Algorithm (GA) (Maulik, U., & Bandyopadhyay, S. 2002; Qu H. et al., 2013), Ant Colony Optimization algorithm (ACO) (Lazarowska A. 2015), etc; Potential field algorithms include Vector Field (Xu H. et al., 2016), Artificial Potential Field (APF) method (Kohei, 2016), and Fast Marching Method (FMM) (Song et al., 2017), etc.

Heuristic algorithms generally consume more time and memory with the increasing of computing range. At present, although there are many applications which need to use Heuristic algorithms, the algorithms cannot adapt to complex environments well. Evolutionary algorithms can adapt to complex environments, but they may often fall into local optimum; the path of calculation often is stochastic, and the computational time is long. Potential field algorithms are widely used in recent years, but it is also easy to fall into local optimum.

For USVs route planning application， since the dynamical constraints are not considered in this stage, in terms of planning behavior, a USV has no difference from a particle. It can reach somewhere without any posture. In addition, as for planning space the application scenarios are usually in large areas, like the open sea. As for planning criteria, the planning approaches for USVs in this stage often combine the classical algorithms mentioned earlier with marine instruments and regulations (as shown in Figure 9) to plan out an optimal route.

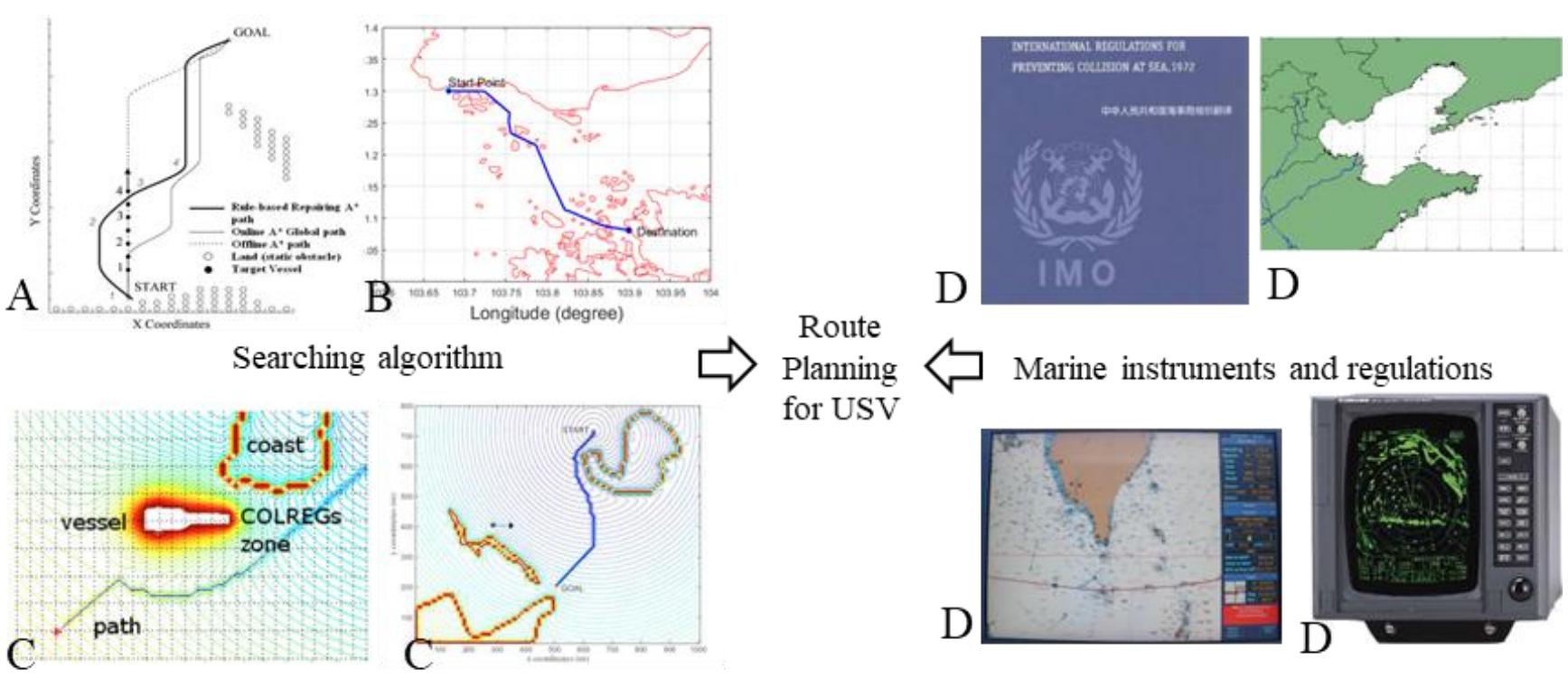

Figure 9 The common way of route planning for USV (A(Campbell, S. & Naeem, W., 2012), B(Niu, H., et al., 2016), C(Naeem, W., 2016), D (Baidu website gallery))

USV Route planning based on the macro-modality constraint, in terms of route-planning space, the searching methods' constraints are the same as environmental constraints, it is not necessary for USVs to consider their specific shape, kinematics, and dynamics.

Jiayuan, et al. (Jiayuan, et al., 2011&2012) combined electronic chart (e-chart) and marine radar with Dijkstra algorithm (Dijkstra, E.W. 1959) to achieve global and local route planning for USVs. They regarded a USV as a particle in the macro-modality environment and used e-charts to get global environment information. And then with the help of the Dijkstra algorithm, they can plan out a global path for the USV. During the voyage, they used a smoothing method to deal with an original radar image and transformed it into a binary image (map) which was used directly for path searching. Finally, with the help of the Dijkstra algorithm, a local path can be planned-out in real-time. Wang Chenbo et al. (Wang Chenbo et al., 2018) realized route planning of a USV in an unknown environment based on reinforcement learning method. This method can achieve better route planning after a certain number of iterations, but it does not consider other environmental factors, such as moving obstacles. Gu, S. et al. (Gu, S. et al., 2019) proposed a route planning method for a USV based on topological location relationship. They constructed the topological map by the topological location relationship, which simplified the search space, and realized the route search of a USV by combining a Dijkstra algorithm. Song, R. et al. (Song, R. et al., 2019) based on the conventional A* algorithm to carry out the USV search path, by reducing the unnecessary nodes to achieve the smooth path, and its practical ship experiments verified that the smoothing A* algorithm can reach the target point in a shorter path.

In terms of route-planning time, planning time is discretization. For example, by making use of e-chart, Jian, L. (Jian, L. 2014) chose the APF method to search a path for a USV in the macro-modality environment. They assumed that there were many small islands and reefs near the shore, thus it is important for a USV to plan out a safe path among these obstacles. Firstly, they marked out the obstacles and feasible zone on an e-chart. And then they improved the repulsive potential to make sure the resultant potential on goal point is the minimum. Finally, they calculated the safest path from the start point to the goal point by using a sequence of decisions.

Lazarowska (Lazarowska, A. 2020) proposed an approach to solving a security problem in routing planning, which utilizes the concept of a discrete APF and a path optimization algorithm, calculates an optimized collision-free route path for a ship. The simulation results show that this method can effectively realize ship route planning in near-real-time.

In terms of route-planning behavior, it has no restriction to research objects' actions. For example, to avoid the collision in the macro-modality constraint environment, the International Marine Collision Regulations (COLREGS) has to be mentioned in the maritime domain (U.S. Department of Homeland Security, 2010). All the ships must obey the rule to prevent collision accident at the sea. The regulations are introduced to normalize the planned path. Researchers (Yanshuang, 2010; Huai Yang, 2016; Campbell, 2012) firstly planned out a global path by A* algorithm in the macro-modality constraint environment. And then based on COLREGS, they classified collisions into different situations and worked out different behavior to avoid each situation. The behavior corresponded to different paths. Naeem, et al. (Naeem, et al., 2016) combined COLREGS with modified APF algorithm to generate COLREGS-compliant trajectories in the presence of both stationary and dynamic obstacles.

In terms of route-planning criteria, it is important to plan out an optimal route generally. Song et al. (Rui Song, et al., 2017) proposed a novel multi-layered fast marching (MFM) path planning method for unmanned surface vehicle path planning in a time-variant maritime environment. The proposed MFM method to provide a safer and more optimal path by using two operation handlers. The path keeps a safe distance away from obstacles and reduces the energy cost by following counter-flow areas. They considered route planning in the context of macro-modality constraints only, and USV dynamics, moving obstacles and sufficient trajectory smoothness have not been considered in the results. Yuxuan, Z., et al (Yuxuan, Z., et al., 2017) surveyed a coverage path planning algorithm, in which they proposed the dynamic raster method with the heuristic search algorithm to achieve a more reasonable and effective path compared with Yang, S. X. & Luo, C.'s method (Yang, S. X & Luo, C. 2004), similarly, they can not consider the kinetics and dynamics of the USV.

## 3.2 SUMMARY OF ROUTE PLANNING

In summary, there are plenty of sophisticated algorithms proposed in this stage, and most of them can perfectly adapt to the application of USV route planning. But Route Planning is not a very accurate path. As it does not consider the dynamic constraints, it is difficult to apply to some actual scenarios that require more precise path planning.

USV Route Planning can be described as follows:

(1) USV Route Planning problem can be transformed into a reachable problem (R).

(2) USV Route Planning constraints condition:

1) Planning Space: The planning space is only constrained by the connectivity of the environment;

2) Planning Time: Planning time is discrete;

3) Planning Behavior: There are no requirements for planning behavior;

4) Planning Criteria: The planning criterion is to solve an optimal route, focusing only on the accessibility, and not on its own constraints and how to get there.

(3) USV Route Planning Model:

$$R = \begin{cases} connected & R = 1 \\ un-connected & R = 0 \end{cases}$$

R=1 means reachable, and R=0 means unreachable. In other words, the route planning problem can be summarized as reachable or not. Because route planning ignores dynamic factors and its own shape, it simply abstracts the planning problem into a line selection problem from one particle to another.

(4) Route planning based on Macro-modality constraint: scope of application, modality problem, constraint problem, and typical approaches are shown in Table 5.

Table 5. Route planning based on Multi-modality constraint: scope of application, modality problem, constraint problem and typical approaches

| Scope of application | Modality problem | Constraint problem | Typical approaches |
|---|---|---|---|
| In the macro-modality constraint environment like RPG (Role-playing game), vehicle navigation in city, indoor navigation for | Macro-modality problem | No kinematics and dynamics constraint problem, research object as a particle | Dijkstra algorithm, A* algorithm, APF, PSO algorithm, GA, ACO algorithm, FMM, etc. |

| robots, and ship routing,<br>etc. |
|---|

# 4. TRAJECTORY PLANNING.

Along with the clues, some scholars begin to pay closer attention to the dynamic constraints of the research object itself. When considering a research object in a physical system, the planning path cannot be as ideal as shown in Figure 5. Hence the study of Trajectory Planning is of vital importance.

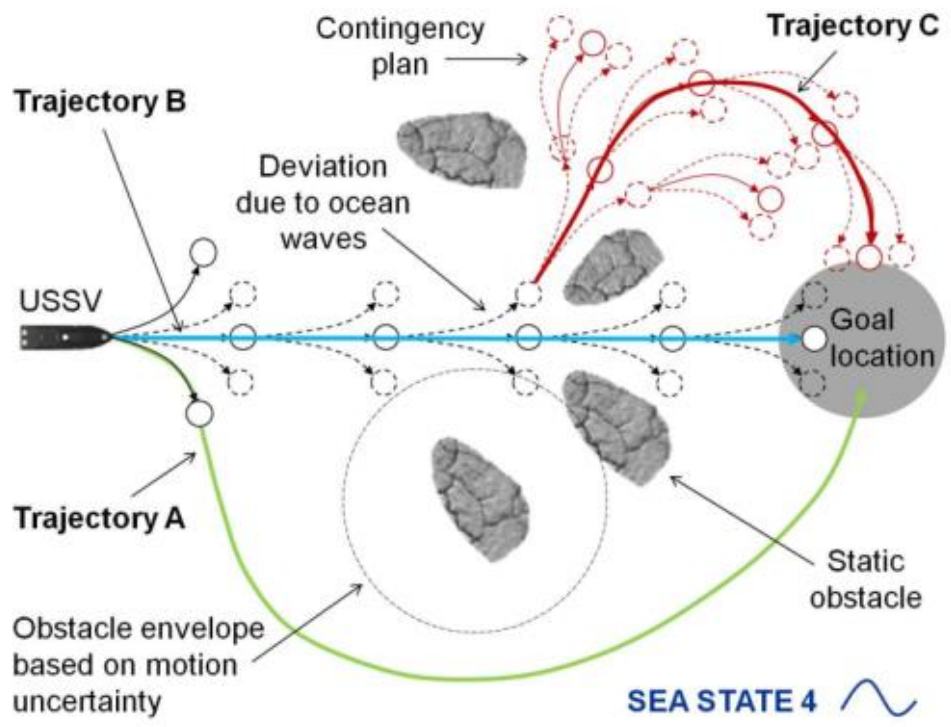

Figure 10 The application in USV Trajectory Planning (Svec, P. et al., 2011)

Trajectory planning can be seen as an improvement of Route Planning. At this stage, Trajectory planning is defined as a medium-modality constraint problem. Instead of regarding a research object as a particle, this stage begins to consider its part dynamic constraints, like size, speed, heading, curvature. In other words, what the Trajectory Planning wants to do is to make the planning path smooth and continuous, and conforms to one or a few of the physical characteristics of the research object (as shown in Figure 10).

## 4.1 TRAJECTORY PLANNING ANALYSIS

The common approaches in this stage are classified into two types, curve fitting and multi-constraint optimization (as shown in Figure 11).

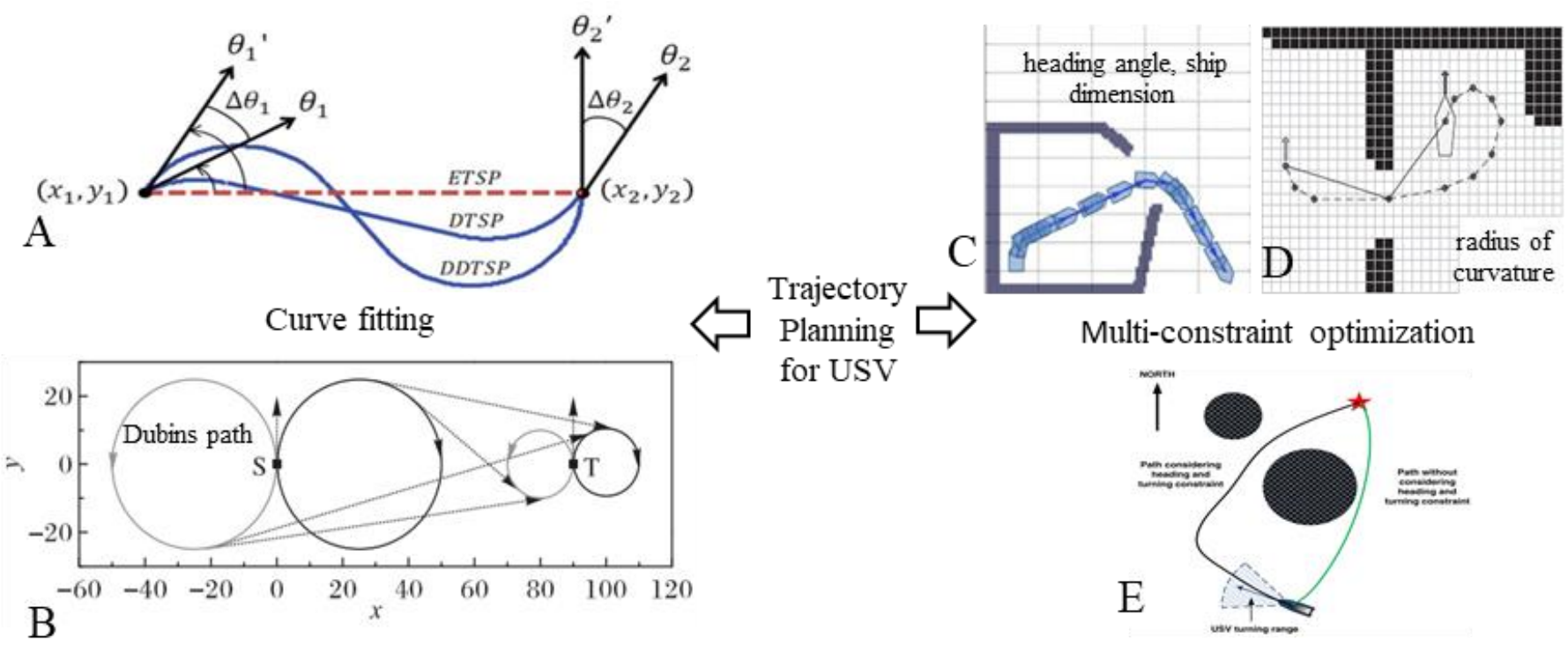

Figure 11 Two approaches of trajectory planning for USV (A (Cohen, L., et al., 2017), B (Liu, L., et al., 2017), C(Yang, J. M., et al., 2015), D(Kim, H., et al., 2014), E(Liu, Y. & Bucknall, R. 2016))

Curve fitting is a way that constructs a curve to make the best fit to the certain polyline. For a USV, it is a way to find the best-fitted curve for the planned path. In terms of trajectory-planning space and behavior, it considers environment conditions and the part of USV dynamic constraints. For example the curve should meet the constraint of the minimum turning radius. It makes the planned path smoother and more continuous in a medium-modality constraint environment. A common approach is a Dubins path (F. Dougherty, et al., 1990). The idea of this method is to find the shortest possible path for a particle with unity speed that meets a minimum curvature bound between a starting pose (which contains position and orientation) and a finishing pose consisting of at most three pieces, and each piece is either a straight line or an arc of a circle. The concrete path composition can be "CLC", "CCL" or "CCC" ("C" and "L" stand for circle and line respectively) (Anderson, et al., 2014).

This method was first applied in UAVs, but in recent years, there are also some applications of Dubins path in USV trajectory planning (Yi Chen, 2016; Liu L. et al., 2017). For example, Yi Chen (Yi Chen, 2016) considered a turning process in the actual sailing of the USV in the medium-modality constraint environment. He combined the "CLC" Dubins path with a genetic algorithm to solve the path optimization problem of the pose (positions and heading angles) transition from the starting point to the end point. However, the Dubins path does not result in curvature continuous paths due to that the curvature of a straight line is 0, whereas a circle arc has a certain curvature. Hence, there will be a jump in the curvature from 0 to a certain value when moving from the straight line to the circle arc. To overcome the shortcoming, some scholars replaced circular arcs to some special curves in the medium-modality constraint environment, like Fermat's spiral (FS) (Dahl A. R., 2013), B splines (Hong Li, et al. 2016) and Bessel curve (Yi Wang, et al. 2012). These curves are good means to connect successive straight lines, because the curvatures of lines are equal to zero at the origin, a property which makes lines suitable for being connected into a straight line without inducing curvature discontinuities. Xiaojie Sun (Xiaojie Sun, 2016) used floating-point numbers and the turning radius of KT equations as the order and the radius of curvature of Bessel curve to restrain the path, and finally got a smooth and continuous trajectory for USVs. Wang, N., et al. (Wang, N., et al., 2019) proposed a multilayer path planning algorithm to search path in complex marine environments. In which they B-Spline method to minimize yaw-cost for the USV path planning.

Multi-constraint optimization is more direct in the medium-modality constraint environment. The main idea is to add part dynamic constraints to the original route planning algorithm. It prompts the planned path more practical and reasonable. For example, Kim, M., et al. (Kim, M., et al., 2014) added a heading angle and radius of curvature constraints to optimize the path. They took into account the possible USV's steering heading angles in the process of planning, turning the original two-dimensional ($x$, $y$) planning space into three-dimensional ($x$, $y$, $\theta$). At the same time, they improved the A* algorithm by adding the radius of curvature into actual cost function g(x) to make turning process more reasonable and effective. In the medium-modality constraint environment, they focused on considering the adequacy of planning space. Liang Hu, et al. (Liang Hu, et al., 2019) proposed a new on-line path planning method for USVs which generates collision-free and COLREGs-compliant paths using a multi-objective optimization approach based on the particle swarm optimization. The proposed method determined the type of encounter and whether a target ship complied with the COLREGs. Therefore, in the medium-modality constraint environment, it is more concerned about the adequacy of planning behavior, such as considering part restrictions in research objects' actions (such as the velocity, heading course, etc.).

In addition, Yang J. M., et al. (Yang, J. M., et al., 2015) considered the ship's initial heading and its size constraints. Firstly, they proposed the Finite A* algorithm by improving A* algorithm to get a shorter path compared with post-A* algorithm (Botea, A. et al., 2004) and A* algorithm, and then they considered the ship's initial heading and size to achieve a safer and smoother path. Others (Zuquan, et al., 2015) focused on the constraint of speed to avoid dynamic obstacles. Firstly they used the Particle Swarm Optimization algorithm (PSO) to plan a path in a static environment. And then they made use of relative speed between the USV and dynamic obstacle to find a navigable path. Moreover, in the medium-modality constraint environment, He, Y., et al. (He, Y., et al., 2017) proposed quantitative analysis of COLREG rules and seamanship for autonomous collision avoidance at open sea. They presented a situation identifying the model and defined four distinct stages in the entire encounter process for autonomous collision avoidance, in which appropriate collision avoidance actions can be taken according to different situations and stages, and they also built quantitative computing models for first time-in-point of the close-quarters situation and first time-in-point of immediate danger. Rong, H., et al. (Rong, H., et al., 2019) investigated the uncertainty of ship trajectory prediction via a probabilistic prediction model. In which they established a parameters of the probabilistic models according to historic ship trajectory data that is from Automatic Identifcation System (AIS). Chen C., et al., (Chen, C., et al., 2019) proposed a trajectory planning algorithm that considered USV kinematics equation in the planning process. In which they utilized the Q-learning (Watkins & Dayan, 1992) and USV features to achieve a better path compared A* algorithm, RRT (Rapid-exploring Random Tree) algorithm. However, they did not consider the dynamic obstacles and ship collision avoidance rules in the planning area, furthermore, the Q-learning algorithm that they used may be too simple and restricted compared other artificial intelligence algorithms such as deep deterministic policy gradient algorithm (Lillicrap, T. P. , et al., 2015) and distributed proximal policy optimization algorithm (Heess, N., et al., 2017).

# 4.2 SUMMARY OF TRAJECTORY PLANNING

In the trajectory planning stage, the planning space, environment conditions and part of dynamic constraints are considered, such as dubins method considering environmental accessibility and shape of an object, etc. As for planning time, it is a sequence of decisions. As for planning behavior, only part restrictions in research objects' actions, such as curve fitting and multi-constraint optimization take into account speed and turning radius of the object, etc. As for planning criteria, it is necessary to plan out an optimal trajectory.

Therefore, Trajectory Planning problem can be described as follows:

(1) USV Trajectory Planning problem can be transformed into a reachable problem (R').

(2) The constraint conditions of USV Trajectory Planning are as follows:

1) Planning Space: The planning space has environmental connectivity constraints and partial self-size and part dynamic constraints;

2) Planning Time: Planning time is discrete;

3) Planning Behavior: there are some planning behavior requirements, such as speed, heading, etc;

4) Planning Criterion: The planning criterion is to optimize the trajectory, focusing on whether the trajectory can meet both the environmental connectivity and some of the USV's static(size, shape, etc.) and part dynamic constraints.

(3) USV Trajectory Planning Model:

$$R' = \begin{cases} part\ of\ constraint \quad connected \qquad R' = 1 \\ part\ of\ constraint \quad un-connected \quad R' = 0 \end{cases}$$

R'=1 means reachable, and R'=0 means unreachable. The problem of trajectory planning can be summarized as the issue of reachability under partial constraints. Because the trajectory planning only considers part of the dynamic and shape constraints, the problem of the trajectory planning can be seen as a problem that a shaped object can reach from one location to another.

(4) Trajectory planning based on Multi-modality constraint involving the scope of application, modality problem, constraint problem, and typical approaches is shown in Table 6.

Table 6. Trajectory planning based on Multi-modality constraint: scope of application, modality problem, constraint problem and typical approaches

| Scope of application | Modality problem | Constraint problem | Typical approaches |
|---|---|---|---|
| In the medium - modality constraint environment USVs, UGVs, UAVs, etc. | Medium-modality problem | Part of dynamic constraint (i.e. shape+kinematics constraint) problems, like size, speed, heading, curvature and etc | Dubins approaches and related improvement approaches, Fermat's spiral, B splines, Bessel curve, PSO algorithm, A* algorithm, Dijkstra algorithm, etc. |

Although trajectory planning makes planning path better and closer to the real trajectory, the part dynamic constraints, in fact, are independent of each other and not full enough. Because those constraints are considered separately, which means one or two of them are being added linearly into the algorithm, but the interactions between them are neglected. In addition, trajectory planning does not really achieve the final goal ("How should I do to get there") of path planning for the USV. How to achieve the final goal will be addressed in the next stage.

## 5. MOTION PLANNING.

Motion Planning is the final stage and the final goal for path planning. At this stage, Motion planning is defined as a micro-modality constraint problem. It can be seen as fine control for a research object, which focuses on whether the planned path can be found via its own control system. To achieve this goal, the dynamic characteristics of the research object should be studied deeply. But unlike the Trajectory Planning which only needs to consider one or two dynamic constraints separately, Motion Planning must treat a research object as a rigid body. That means all the dynamic constraints should be considered. Thus, it is important to study the mathematical model in this stage.

## 5.1 TWO SYSTEMS OF MOTION PLANNING

According to the control ability, there are two kinds of research objects in this stage: the object of fully actuated system, like the industrial robot and mechanical arm; and the object of underactuated system, like an USV (Figure 12 shows an application in USV Motion Planning).

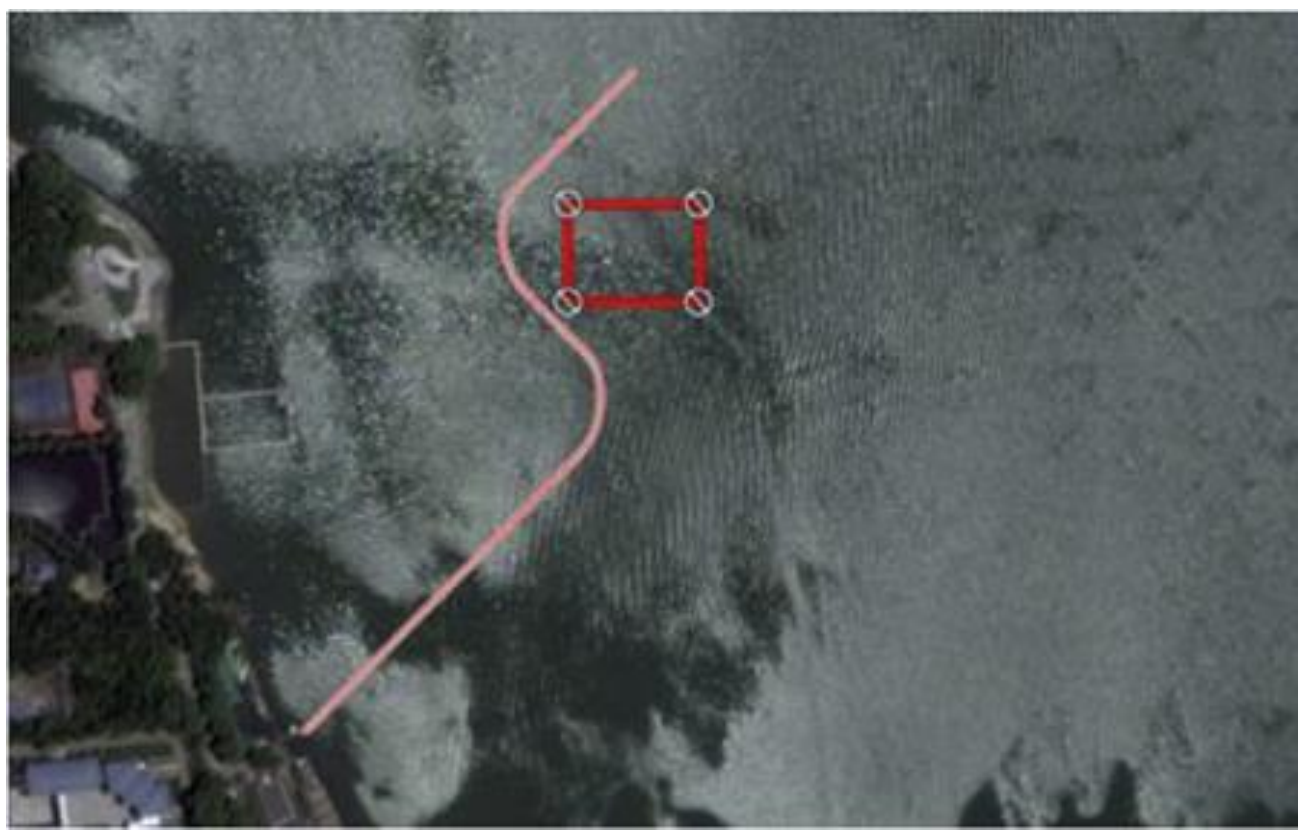

Figure 12 The application in USV Motion Planning (Zhe Du et al., 2018)

For a fully actuated system, the degree of control is equal to the degree of freedom. All the motion poses can be controlled precisely by dynamic equations. As to Motion Planning, the only thing needed to do is to adjust and optimize the control parameters to realize different paths for different tasks. Thus, it is easy to deal with the problem when the research objects are fully actuated system. However, for an underactuated system, because the degree of control of the underactuated system is less than the degree of freedom, not all the motion state can be achieved. Thus, for an underactuated system, one of the important tasks of motion planning is to figure out which state the research object can reach.

## 5.2 TWO TYPES OF MOTION PLANNING

In general, there are two kinds of methods to solve motion planning problems for the underactuated systems such as USVs. One is based on Control Theory, and the other is based on random sampling.

### 5.2.1 BASED ON CONTROL THEORY

The first one is based on Control Theory. The approach is divided into two steps in the micro-modality constraint environment. Firstly, by making use of traditional path planning (route planning) algorithm, a collision-free route from the start point to the end point is planned out. Then, based on the mathematical model of a research object, a controller is designed, which contains kinematics and dynamics constraints. Taking advantage of the motion controller, the final path is produced by the second planning (Liang Kang, 2010).

The key point of this approach is motion control that uses control strategy to reconstruct or approximate the "ideal path" which is planned for the first time. With different goals, the motion control can be classified into point stabilization, path tracking and path following (Hao Yu. 2014). Although this approach can be used to solve the problem to a certain degree, the way of solving belongs to the field of control rather than the field of planning in the micro-modality constraint environment.

For example, Escario et al. (Jose B. Escario, et al., 2012) studied the optimization of USV motion planning by applying Ant Colony Optimization metaheuristic. The aim of their work is to find optimal maneuvers for an USV by using an ant algorithm. They modified the original ACO algorithm and the algorithm gave an optimal path in the micro-modality constraint environment. Ma et al. (Ma, Y., et al., 2018) studied the multi-objective path planning problem for USVs in environments with currents effects. They focused on path planning problem that tends to simultaneously optimize four objectives (the path length, the path smoothness, the economic cost, and the path safety) and at the same time it is subject to the collision avoidance, motion boundaries and currents effects in the micro-modality constraint environment. It takes dynamic behavior into account and gives an optimal motion planning.

Moreover, Zhe Du et al. (Zhe Du et al., 2018) established trajectory units based on ship mathematics which is Maneuvering Mathematical Group (MMG) model (Ogawa, O., 1977), and considered the dynamics requirements of a USV, and then proposed a motion planning algorithm for an USV. Based on the literature (Zhe Du et al., 2018), Zhe Du et al. (Zhe Du et al., 2019) further proposed the 32 Directional USV trajectory units and a new path search algorithm for the USV motion planning. Bitar, G., et al. (Bitar, G., et al., 2019) considered a warm-started optimized motion planning algorithm for USVs. Firstly, they harnessed A* algorithm to get shortest piecewise linear path, but the kind of path is not feasible for USV motion planning; Secondly, they connected the waypoints that are acquired by A* algorithm using circles and adds artificial dynamics, to make the path become closer to feasible and practical path. However, they may lose the optimality when modifying the shortest waypoints.

### 5.2.2 BASED ON RANDOM SAMPLING

Another approach is based on Random Sampling in the micro-modality constraint environment. All the methods in Route Planning stage plan out a deterministic result. That is to say, in the same planning space, the results of any two kinds of planning are the same. However, the dimension (constraints) of the motion planning is much higher than that of route planning, and with the dimension increasing the complexity of the planning problem grows exponentially (NP-hard problem (Canny, 1998)). Some scholars proposed Random Sampling based on motion planning algorithms to solve the problems in the micro-modality constraint environment. The biggest characteristic of these methods results in that any two kinds of planning are different, even in the same planning space. Probabilistic Roadmap Method (PRM) and Rapidly-exploring Random Tree (RRT) are the typical ones.

Firstly, PRM algorithm is proposed by KavrakiL et al. (KavrakiL, et al., 1994) and the algorithm considers all restrictions in research objects' actions. The main idea is to generate random road signs (state) in the space to determine the feasible region (free space), then connect the adjacent road signs to become random roadmap, and finally find out a feasible path on this random roadmap. The nodes on the map represent the pose of the research object, and the connecting lines between those nodes (the edges in the graph) represent the feasible path between different poses. Similar to this idea, the State Lattice (M. Pivtoraiko et al., 2005) is a discretization from the configuration space to a set of states, representing configurations, and connections between these states, where every connection represents a feasible path.

To move a step forward, RRT algorithm is more applicable to an optimal control-related problem and a nonholonomic system (Nan, 2014), and the algorithm considers all the dynamic constraints in the micro-modality constraint environment. The main idea is that by corresponding input of control, randomly generates a search tree from the initial state. Each vertex represents a state and each directed edge represents a control input. With the tree growing, the state updates constantly. When a vertex reaches the desired target area, the tree stops growing (S. M. LaValle, 1998). Because the state transition in RRT algorithm can be improved by kinetic models and new node generation can be constrained by kinetic equations, it can be combined with the mathematical model to solve motion planning issues in the micro-modality constraint environment (Na, et al., 2011; Mingbo, et al., 2015; Jinze Song et al., 2010). A popular research direction currently is to apply the RRT algorithm to the dynamic environment. The guidance factor is introduced to the direction of node expansion to enhance the purpose of the search tree's growing (Guangzhen et al., 2013; Heoncheol et al., 2012).

Approaches based on Random Sampling are widely used in motion planning of the robotics (such as industrial robots, UAVs, etc.). However, like inertia, resistance and response time on waters are larger than those on the ground, the motion control for USVs is far more complex and difficult than UGVs or UAVs, and the above methods of motion planning need to be improved for USVs motion planning. Moreover, with the increasing density of water traffic, the scope of the sailing area is decreasing gradually, and where the USVs need to be more precisely steered.

## 5.3 SUMMARY OF MOTION PLANNING

In the motion planning stage, the planning space, environment conditions and dynamic constraints are considered. For example, control theory and random sampling consider environmental accessibility, the shape of object, kinematics and dynamics of an object, etc. As for planning time, it is a sequence of decisions. As for planning behavior, restrictions in research objects' actions, such as speed, distance, moment and inertia of the object, etc. are considered. As for planning criterion, it is normally necessary to plan out an optimal motion planning.

Therefore, the Motion Planning problem can be described as follows:

(1) USV Motion Planning problem can be transformed into a reachable problem (R'').

(2) USV Motion Planning constraints condition:

1) Planning Space: The planning space has environmental connectivity constraints and its own size and dynamic constraints;

2) Planning Time: Planning time is discrete;

3) Planning Behavior: Planning behavior requires comprehensiveness, not only to pay attention to when to give more power but also to pay attention to when and how many rudder angles are appropriate;

4) Planning Criterion: The planning criterion is to solve a path containing specific behavioral instructions, focusing on how to navigate to reach the target point, not only depending on the environmental connectivity, but also on how to navigate under the USV dynamics constraints and their own size constraints.

(3) USV Motion Planning Model:

$$R'' = \begin{cases} full\ of\ constraint \quad connected \quad R'' = 1 \\ full\ of\ constraint \quad un-connected \quad R'' = 0 \end{cases}$$

R'' = 1 means reachable, and R'' = 0 means unreachable. The problem of motion planning can be summarized as the problem of reachable under the constraints of dynamics and its own shape and size etc. Because motion planning considers dynamics and shape size constraints, etc., the USV motion planning is considered as an object with size constraints and dynamic constraints, and how to operate from one location to another location.

(4) Motion planning based on Multi-modality constraint including the scope of application, modality problem, constraint problem, and typical approaches is shown in Table 7.

Table 7. Motion planning based on Multi-modality constraint: scope of application, modality problem, constraint problem, and typical approaches

| Scope of application | Modality problem | Constraint problem | Typical approaches |
|---|---|---|---|
| In the micro-modality constraint environment the industrial robot, mechanical arm, UGV, UAV, USV and etc | Micro-modality problem | Fully dynamic constraint (i.e. shape+ kinematics +Dynamics constraint) problem | A* algorithm , PRM algorithm, dijkstra algorithm, PSO algorithm, ACO algorithm, RRT algorithm etc. |

In some special navigation environment, a small range of autonomous control and fine operation becomes necessary, such as the task of self-avoidance and self-berthing in harbor areas. Svec et al. (P. Svec, et al., 2012) have made some attempts. They took advantage of the probability prediction method to solve the problem. They first used a USV mathematical model to predict all the possible trajectories, and then employed min-max game-tree search to select the trajectory which had the smallest collision probability from the possible trajectories pool. But it is still unclear how these possible trajectories are generated (or what the special properties they have) and how to control an USV (the specific instruction) to achieve the planned path.

# 6. CONCLUSION AND OUTLOOK.

This paper focuses on path planning for USV based on the multi-modality constraint. With the research development as a clue, the paper introduced the common approaches and algorithms in three stages: Route Planning, Trajectory Planning and Motion Planning.

(1) Route Planning is a basic stage and Route Planning problem is a macro-modality constraint problem. It treats a USV as a particle. So the approaches in this stage often combine the classic algorithms used in USV areas with marine instrument and regulations.

After the above-mentioned summary of the existing technologies of USV route planning based on macro-modality constraint environment, the combination of navigation rules, semantic rules and route planning are rarely studied. Those ways could help achieve smarter USVs.

(2) Trajectory Planning is a transition stage and Trajectory Planning problem is a medium-modality constraint problem. It considers the part dynamic constraints (such shape and kinematics) of USVs. The methods in the stage usually have two types: one is curve fitting and the other is multi-constraint optimization. Curve fitting is a way to construct a curve to make the best fit for the planned route path (in route planning). Multi-constraint optimization is more direct in that it improves the original planning algorithm by adding kinematic constraints to make the final path to meet these constraints. The former makes the planned path smoother, while the latter makes it more reasonable.

After the above-mentioned summary of the existing technology of USV trajectory planning based on medium-modality constraint environment, there are still some things to do to realize real-time trajectory planning.

(3) Motion Planning is the final stage and Motion Planning problem is a micro-modality constraint problem. Motion Planning is the ultimate goal of path planning for USVs. It treats an USV as a rigid body and focuses on motion control. Thus the approaches in this stage link with control theory and motion prediction. But there is no effective way to guide a USV how to command the control systems to make it get to the goal all by itself. Therefore, there is still much work to do.

According to the above summary of the existing technologies of USV motion planning based on micro-modality constraint environment, full autonomous USV motion planning has not yet been achieved.

In addition, the main challenges of path planning are different, because planning demands in each stage are different.

As for an USV or ship's routing planning, its application scenarios are often a macro-modality constraint in large scale area. For example, it is a path planning from a port to another port, of which the main challenges are weather change, tsunami and typhoon, etc.

As for an USV or ship's trajectory planning, its application scenarios are often medium -modality constraint in medium scale area. For example, the path planning is how to navigate a ship into a port, of which the main challenges are how to avoid the obstacles, including static obstacles and dynamic obstacles.

As for an USV or ship's motion planning, its application scenarios are often precise operations which need micro-modality constraint in small scale area. For example, the path planning is how to maneuver a ship to berth alongside, of which the main challenges are how to steer stably and avoid wind, wave and current effects, furthermore, to prevent the ship's drift which has a negative impact on ship's tasks.

Therefore, looking forward to the future development of USVs path planning, we need to proceed from the following directions to help achieve smarter and safer USVs: the multi-element integration of perception system, steady and robust control system, intelligent decision-making system, diversified task execution and complex environment adaptation.

In general, the development of path planning for USVs, which still needs to combine with interdisciplinary approaches to solve problems such as the berthing problem and the path-planning problems for USVs under complex sea conditions (wind, wave, current and weather interference, etc.), which still need to be further solved. In order to solve the above problems, it is necessary not only to consider the actual hydrodynamic influence of USVs, but also to find a reasonable solution to the N-P difficult problem (Michael RG and Johnson DS, 1979; Liu, S. et al., 2011; Ma, Y. et al., 2018). For example, the planned path should consider the purposes that are the smoothness of rudder, the distance of navigation, the economy of navigation and the safety of navigation, etc.

In conclusion, for the short-term and long-term goals of the development of USVs path planning, we should start from the following directions:

1. Short-term objectives:

(1) Quantitative rules and path planning.

The navigation of the USV needs to conform to the navigation rules, and the rules should be quantified to further make the navigation of the USV conform to the collision avoidance rules well. At present, some scholars have made some achievements in this field, such as Zhuang, J. Y., et al. (Zhuang, J. Y., et al., 2011) and He, Y., et al. (He, Y., et al., 2017) etc.

(2) Semantic situation and path planning.

In the future, it is necessary to research the effect of space semantic situation (Zhang, Y., et al., 2018; Wen, Y., et al., 2019) on USV path planning. At present, semantic planning and control are rarely seen in the path planning of USVs, but some scholars have made some research in the mission planning of unmanned underwater vehicles (Patron, P. 2010) and path planning of ground mobile robots (Wang, C. et al., 2019; Yaagoubi, R., et al., 2020). Because the semantic environment can greatly reduce the reliance on complex multi-source data, and greatly reduce the amount of calculation, and then improve the efficiency of calculation.

(3) Advanced control methods and path planning.

The navigation process of USVs is a nonlinear control and uncertain process, the navigation control of USVs in the future needs to combine with advanced control methods. Generally, the control theory is far from practical application, especially there are fewer advanced theoretical control methods that are applied for USVs, such as nonlinear control based on manifold space (Sakamoto, N. 2013) and partial differential equations control (Zheng, J. & Zhu, G. 2019) and so on.

(4) Path planning under complex sea conditions.

It is necessary to propose methods which plan the route and motion commands for USVs in real time under complex sea conditions. In this respect, some scholars have made some researches, such as Song, R., et al. (Song, R., et al., 2017), Ma, Y., et al. (Ma, Y., et al., 2018) and Wang, N., et al. (Wang, N., et al., 2019), etc.

2. Long-term objectives:

(1) USVs autonomous berthing.

At present, the intelligent level of USVs has been improved. However, as for autonomous berthing, especially under the influence of current, wind and shore wave effect, is still difficult to achieve. Therefore, it is necessary to research the effects of current, wind and shore wave on autonomous berthing. Some scholars have done some research on this topic, such as Djouani & Hamam (Djouani, K., & Hamam, Y. 1995),Okazaki, T., et al. (Okazaki, T., et al., 2000), Meyer, P. J., et al. (Meyer, P. J., et al., 2019) and Martinsen, A. B., et al. (Martinsen, A. B., et al., 2019).

(2) USVs formation coordinated motion planning control.

It is an inevitable trend for the coordinated motion planning control of USVs in the future, because it is difficult to complete complex tasks by relying on a single USV, and the research on the coordinated control of multiple USVs is a key point for the development of USVs formation in the future. Liu, B. et al. (Liu, B. et al., 2019), Chen, L., et al. (Chen, L., et al., 2019) and Li, S., et al. (Li, S., et al., 2019) etc. have done some researches in this area.

(3) Realization of group USVs autonomous navigation in complex environment.

At present, the motion planning and cooperative control of the USV formation are generally carried out in a relatively simple environment, without enough consideration of complex environment, such as strong wind, strong current, waves, multiple moving obstacles, etc., so how to ensure the USV formation to carry out adaptive, efficient and safe navigation task is still a problem to be solved.

(4) USV formation, UAV formation and submarine formation cooperate with motion planning and mission planning control to achieve multi-USVs, multi-UAVs and multi-submarines cooperation to perform complex target tasks.

It is very helpful for the implementation of complex tasks in the future to research multiple robotics and multiple formations for air-sea cooperative motion planning and task planning. For example, in order to deal with oil pollution at sea, it is necessary to monitor oil diffusion in the air and under the sea by multiple UAVs and submarines respectively, and to clean up oil pollution by USV formation. Hence, we should develop the air-sea cooperative mission, especially the motion control and collaborative planning of a variety of UAVs, USVs and submarines, and establish a variety of vehicle control systems.

## ACKNOWLEDGEMENT

This work is supported by the National Key R&D Program of China (No. 2018YFC1407405, 2018YFC0213904); the National Natural Science Foundation of China (No. 51679180, 51579204, 41801375 , 51709218); the Funding of the State Key Laboratory of Surveying, Mapping, Remote Sensing and Information Engineering of Wuhan University (No. 17I03); the Natural Science Foundation of Hubei Province (No. 2016CFB362). The authors are grateful to the three anonymous reviewers for their constructive comments.